\documentclass[letterpaper, 10pt, conference]{ieeeconf}  % Comment this line out if you need a4paper

\IEEEoverridecommandlockouts                              % This command is only needed if 
\usepackage{graphics} % for pdf, bitmapped graphics files
\usepackage{epsfig} % for postscript graphics files
\usepackage{mathptmx} % assumes new font selection scheme installed
\usepackage{times} % assumes new font selection scheme installed
\usepackage{amsmath} % assumes amsmath package installed
\usepackage{amssymb}  % assumes amsmath package installed
\usepackage{amsmath,amssymb,amsfonts}
\usepackage{mathtools}
\usepackage{graphicx}
\usepackage{booktabs, tabularx, array}
\usepackage{pifont}
\let\labelindent\relax
\usepackage{enumitem}
\usepackage[table]{xcolor}
\definecolor{headergray}{HTML}{F3F4F6}
\definecolor{best}{HTML}{E6F4EA}    % 最优：淡绿
\definecolor{second}{HTML}{FFF4CC}  % 次优：淡黄
\definecolor{oursrow}{HTML}{EAE7FF} % ours 行：淡紫
\usepackage{multirow}
\usepackage[colorlinks=true, linkcolor=blue, urlcolor=blue]{hyperref}

\newcommand{\best}[1]{\cellcolor{best}\textbf{#1}}
\newcommand{\secondbest}[1]{\cellcolor{second}#1}
\usepackage{bm}
\newcommand{\vx}{\mathbf{x}}
\usepackage{booktabs}
\usepackage{algorithm}
\usepackage[noend]{algpseudocode}
\numberwithin{equation}{section}

\newcommand{\RR}{\mathbb{R}}
\newcommand{\vc}{\mathbf{c}}
\newcommand{\veps}{\bm{\epsilon}}
\newcommand{\veta}{\bm{\eta}}
\newcommand{\vI}{\mathbf{I}}
\newcommand{\baralpha}{\bar{\alpha}}

\newcommand{\pos}{\mathrm{pos3}} % select the first 3 EE dims (position)

\usepackage{cite}

\usepackage{siunitx}
\title{\LARGE \bf
FORGE-Tree: Diffusion-Forcing Tree Search for Long-Horizon Robot Manipulation
}

\author{Yanjia Huang$^{1*}$, Shuo Liu$^{2*}$,  Sheng Liu$^{3}$, Qingxiao Xu$^{1}$,  Mingyang Wu$^{1}$, Xiangbo Gao$^{1}$ and Zhengzhong Tu$^{1}$    % <-this % stops a space
\thanks{$^{1}$Department of Computer Science and Engineering, Texas A\&M University
        {\tt\small }}%
\thanks{$^{2}$Department of Electrical \& Computer Engineering, University of Washington
        {\tt\small }}%
\thanks{$^{3}$Karlsruhe Institute of Technology, Germany
        {\tt\small }}
\thanks{*Equal Contributors}
\thanks{Corresponding Author: tzz@tamu.edu}
}
\begin{document}

\maketitle
\thispagestyle{empty}
\pagestyle{empty}

%%%%%%%%%%%%%%%%%%%%%%%%%%%%%%%%%%%%%%%%%%%%%%%%%%%%%%%%%%%%%%%%%%%%%%%%%%%%%%%%
\begin{abstract}

\textit{Long-horizon robot manipulation tasks remain challenging for Vision-Language-Action (VLA) policies due to drift and exposure bias, often denoise the entire trajectory with fixed hyperparameters, causing small geometric errors to compound across stages and offering no mechanism to allocate extra test-time compute where clearances are tight. To address these challenges, we introduce \textbf{FORGE-Tree},  a plug-in control layer that couples a stage-aligned \emph{Diffusion Forcing} (DF) head with test-time \emph{Monte Carlo Tree Diffusion} (MCTD). With a frozen VLA encoder, DF aligns timesteps to subtask stages; during inference we partially denoise only a target segment while keeping other tokens frozen, turning trajectory refinement into a sequence of local edits. We then apply Monte Carlo Tree Diffusion to select the next segment to refine. A scene graph supplies priors for expansion and geometry relation-aware scoring for rollouts, yielding \emph{tree-structured denoising} whose performance scales with search budget while preserving the executed prefix. Evaluation on LIBERO, FORGE-Tree improves success rate by \textbf{+13.4–+17.2} pp over the native VLA baselines with both OpenVLA and Octo-Base. Gains remain consistent under comparable compute budgets, especially on long-horizon variants. Videos available at: \url{https://taco-group.github.io/FORGE-Tree/}}

% Addressing the challenge of long-horizon robotic manipulation, existing approaches often struggle to balance generative flexibility with precise, constraint-driven planning. We propose a novel framework that fuses a Diffusion Policy with Monte Carlo Tree Search (MCTS), leveraging the generative model itself to guide and enhance the search process at test time. This integration enables both expressive trajectory generation and adherence to task-specific constraints. Our core contribution lies in embedding a noise-strength-conditioned diffusion head, guided by a vision-language model (VLM), directly into the MCTS loop at inference time. Specifically: 1) We introduce a controllable denoising mechanism that enables MCTS to apply fine-grained control over trajectory generation through independently scheduled sub-task noise levels. 2) We leverage a DDIM-based scheduler to efficiently complete trajectories from partially denoised states, facilitating fast and accurate reward evaluation in 3D space. 3) The use of a VLM for both scene parsing and high-level action proposal ensures semantic consistency throughout the planning process.
% This tight coupling between generative modeling and structured search produces high-quality, globally coherent, and executable robot action trajectories. Our work marks the first deep, bidirectional integration of Diffusion Policy and MCTS, opening new directions for complex, long-horizon robotic planning.
\end{abstract}

\section{Introduction}

Long-horizon, language-conditioned manipulation is a task where success hinges on \emph{global plan coherence} across multiple sub-tasks and \emph{local geometric precision} at stage ends and the final goal \cite{black2024pi_0,kim2024openvla,mu2023embodiedgpt,mu2024robotwin,zawalski2024robotic}. A simple instruction such as \emph{“put both the alphabet soup and the tomato sauce in the basket”} demands stage-wise reasoning (reach–grasp–place, twice), contact-aware motion, and centimeter-level alignment with relational constraints (e.g., \emph{in-basket}). Beyond isolated pick-and-place skills, such instructions require maintaining a consistent task narrative across partial successes and minor slips, while remaining precise enough to satisfy relation constraints that are only locally verifiable. These gaps suggest treating control not as a one-shot commitment but as an explicitly staged refinement process and reallocating attention and computation where uncertainty concentrates.

Existing methods such as Vision–Language–Action models  (VLAs) learn to map images and text to actions \cite{team2024octo,kim2024openvla}, and diffusion policies provide robust multimodal sequence generation \cite{chi2023diffusion, huang2025pandoradiffusionpolicylearning}. However, long horizons expose three recurring gaps. \emph{First}, most decoders denoise entire trajectories with a fixed schedule, so small pose errors compound across stages. \emph{Second}, decoding knobs (e.g., step schedule, guidance, temperature) are typically frozen \emph{a priori}, providing no mechanism to spend additional test-time compute where clearances are tight or the scene is ambiguous. \emph{Third}, feedforward controllers seldom leverage \emph{symbolic or relational} structure alongside continuous kinematics, limiting reliability on relation-heavy tasks.

To address these issues, we keep the VLA encoder fixed and upgrade only the control layer with \textbf{FORGE-Tree}. The key idea is to treat diffusion decoding not as a one-shot generator but as a \emph{planner over partial trajectories}. This perspective decouples planning from perception-scale retraining and turns decoding into a sequence of transparent, budget-aware edits over short horizons. Concretely, FORGE-Tree (i) aligns denoising to stage structure so the model \emph{learns to land} subgoals, (ii) casts inference as a budgeted search over \emph{denoising meta-actions}, which segment to edit next and how aggressively to denoise it, so compute is adaptively focused where it matters, and (iii) couples a lightweight scene graph to both exploration and evaluation, linking symbolic relations (e.g., \texttt{in}, \texttt{on top of}) with continuous robot kinematics. Crucially, these priors act as soft guidance rather than hard constraints, preserving flexibility while exposing the reasons behind each refinement. 
% The result is \emph{tree-structured denoising}: a receding-horizon process that preserves good prefixes, refines short future segments, and scales in performance with test-time budget.
% : a \emph{Diffusion Forcing} head that aligns denoising with stage structure, and a test-time \emph{Monte Carlo Tree Diffusion} planner that turns decoding into \emph{tree-structured denoising}. The key novelty is to search over \emph{denoising meta-actions}, which segment to edit, how jumpy to denoise, how strongly to guide, and how stochastic to sample, so the planner selectively refines short future segments while preserving a good prefix. A scene graph provides \emph{both} exploration priors and geometry or relation aware scoring, coupling symbolic goals to continuous action sequences.
Trained on EMMA-X and evaluated in simulation on LIBERO and ManiSkill under standard protocols \cite{sun2024emma,liu2023libero,mu2021maniskill}, our method improves over a VLA to Diffusion Policy baseline. 
% DF alone reduces mid-horizon drift, and MCTD delivers monotonic gains with increasing search budget, yielding higher success rate, lower terminal pose error and collisions, and higher relation satisfaction on both OpenVLA and Octo backbones.

We position \textbf{FORGE-Tree} as a control-layer upgrade rather than a new pretraining recipe, prioritizing modularity, interpretability, and test-time efficiency. \textbf{Contributions} are as follows: \underline{(1)} A VLA-conditioned \emph{Diffusion Forcing} scheme that teaches the head to \emph{land} stage subgoals. \underline{(2)} \emph{Tree-structured denoising} through meta-actions on segment, stride, guidance, and temperature: scalable adaptive budget decoding. \underline{(3)} A \emph{dual-role scene graph} providing both expansion priors and geometry relation-aware evaluation to bridge symbols and kinematics, making decisions traceable to human-readable relations. \underline{(4)} A \emph{plug-and-play} control layer that drops into VLAs and produces consistent gains without modifying the backbone, easing adoption across existing VLA stacks.

\begin{figure*}[t]
  \centering
  \includegraphics[width=\textwidth,keepaspectratio,pagebox=cropbox]{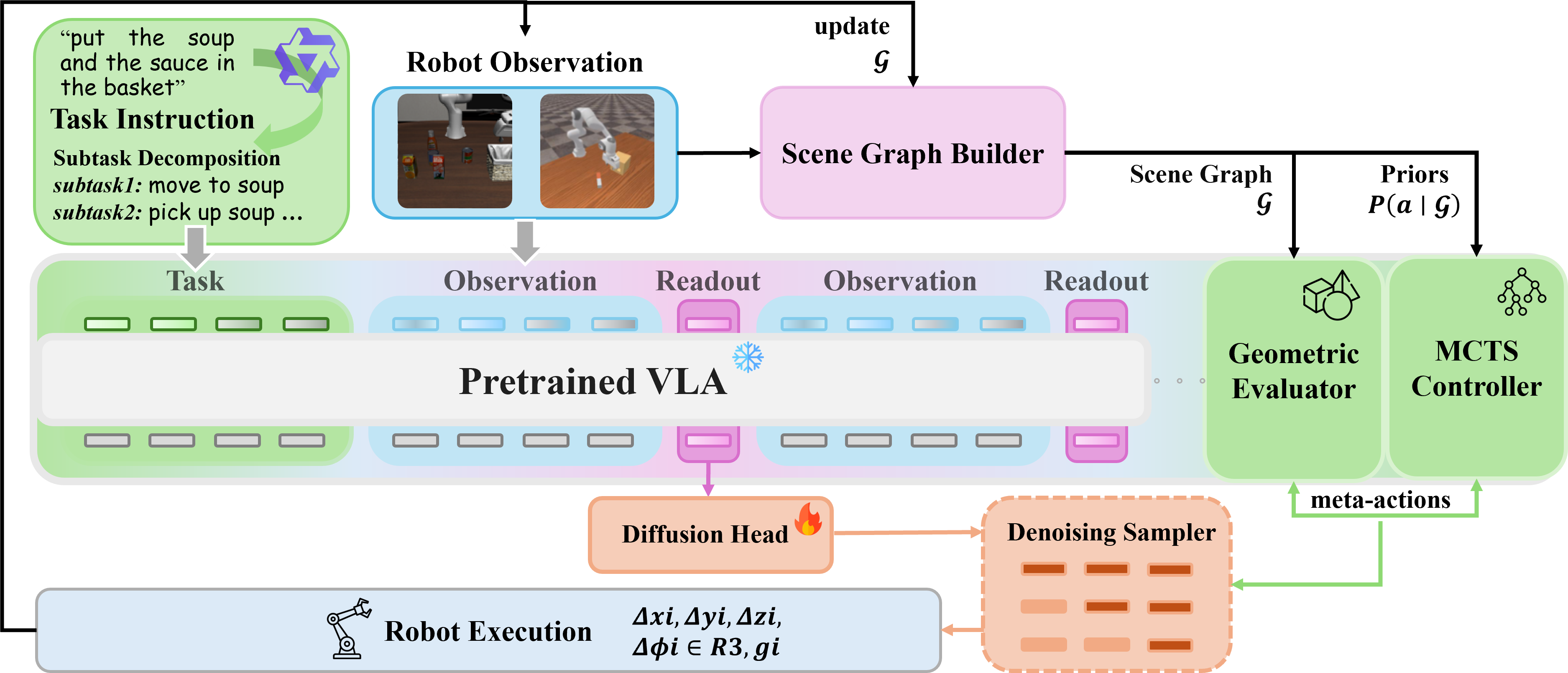}
  \caption{\textbf{System overview.} Pretrained VLAs encode the instruction and observations,
  builds a scene graph $\mathcal{G}$, our diffusion head predicts noise, the partial-denoising
  sampler edits a selected future segment with meta-actions $a=(k,m,s,w,\tau)$, and MCTD evaluates
  candidates with geometry-aware rewards before executing the first segment and replanning.}
  \label{fig:overview}
\end{figure*}

\section{Related Work}
\subsection{Vision-Language-Action Models}
Recent advances in Large Language Models and Vision Language Models have shown that transformer models trained on internet-scale data possess remarkable capabilities in text generation, multimodal content understanding, and reasoning\cite{achiam2023gpt, touvron2023llama, team2023gemini} . Inspired by these developments, researchers have explored using transformer models to directly map language instructions and visual observations to physical robot actions\cite{brohan2022rt, zitkovich2023rt, team2024octo, kim2024openvla, black2024pi_0, lee2025molmoact}. These models, trained on large-scale robotics datasets\cite{khazatsky2024droid, o2024open, lee2025molmoact} via behavior cloning, can generalize to unseen scenes and tasks beyond their training data distribution. However, their autoregressive architecture—designed to predict the next action or action chunk\cite{zhao2023learning}—struggles with cumulative errors and typically fails in long-horizon tasks requiring sequential operations. Our approach builds a test-time framework that enhances pre-trained VLA models with Monte Carlo Tree Search\cite{silver2017mastering} and scene graphs\cite{li2024scene}, leveraging structured search to enable effective long-term planning.

\subsection{Diffusion Models for Robotics}
Diffusion models are generative models that learn data distributions through iterative noise addition and denoising processes, showing excellent results in image and video generation\cite{ho2020denoising, rombach2022high, liu2024sora}. Building on this success, researchers have applied these models to robotics\cite{chi2023diffusion, ze20243d}, where they demonstrate powerful capabilities in multimodal action generation and planning\cite{janner2022planning, mishra2023generative}. Some approaches combine VLM backbones with diffusion heads to generate smooth motion trajectories and use multimodal conditions to guide the denoising process\cite{liu2024rdt, wen2024diffusion, wen2025tinyvla, zhu2025scaling}. In our work, inspired by the Diffusion Forcing mechanism\cite{chen2024diffusion}, we introduce a diffusion action head conditioned on VLA readout information. This is combined with search mechanisms to schedule the denoising process, integrating next token prediction with the diffusion model's denoising time-steps, resulting in more flexible control and improved geometric alignment.
\subsection{Long-horizon manipulation tasks}
Long-horizon tasks remain a persistent challenge in robot manipulation. Traditional approaches like Task and Motion Planning (TAMP)\cite{garrett2021integrated} and skill chaining\cite{mishra2023generative, mishra2024generative} require perfect perception of world states and precise dynamics modeling, which can rarely met in real-world environments. Recently, researchers have leveraged LLMs and VLMs as planners for open-world robot manipulation due to their powerful prior knowledge\cite{rana2023sayplan, huang2023voxposer, duan2024manipulate, ahn2022can}. Yet these approaches still require integration with motion planners to achieve physical robot control. Most similar to our approach is VLAPS\cite{neary2025improving}, they use MCTS and world model with pretrained VLA models to handle long-horizon tasks. Different from this, our method combines VLA models' ability to directly map instructions and observations to actions with VLMs and scene graphs serving as abstract world state representations. We integrate these components with search algorithms to effectively solve long-horizon operation problems.

\section{Method}

% \begin{figure*}[!t]
%   \centering

%   \includegraphics[width=\linewidth]{DF.png}
%   \caption{\textbf{Diffusion Forcing (DF) with partial denoising.}
%   Left: \emph{Noise as masking}—subtasks $\mathcal{S}_j$ share per-subtask timesteps $t[i]{=}t_j$ (darker cells indicate larger $t$).
%   Middle (training): conditioned on $c\!=\!f_{\text{VLA}}(o,u)$, the diffusion head predicts $\epsilon_\theta(x_t[i],c,t[i])$, recovers $\hat{x}_0[i]$, and is supervised by the DF loss (noise MSE + end-of-trajectory and stage-end geometric terms + smoothness) \cite{chi2023diffusion,chen2024diffusion}.
%   Right (inference): we \emph{partially denoise} a selected segment $S_{k,m}$ using jumpy DDIM \cite{song2020denoising} with geometry guidance $w\nabla_{x_t}U(\hat{x}_0;\mathcal{G})$; only the segment evolves while the complement is frozen, and the first segment is committed before replanning.}
%   \label{fig:df}
% \end{figure*}
\begin{figure*}[!t]
  \centering

  \includegraphics[width=\linewidth]{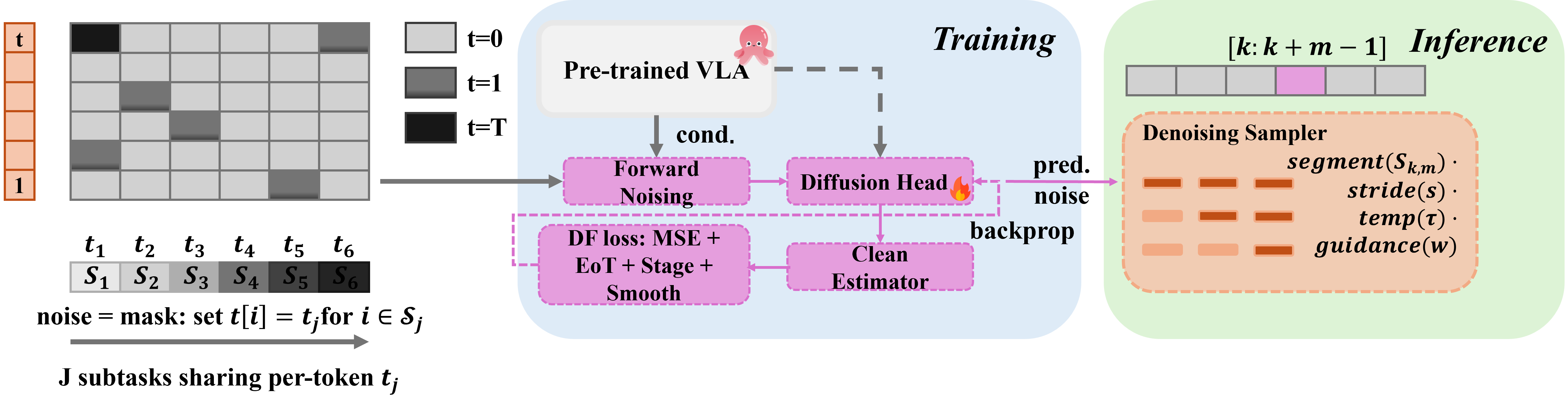}
  \caption{\textbf{Diffusion Forcing (DF) with partial denoising.}
  Left: \emph{Noise as masking}—subtasks $\mathcal{S}_j$ share per-subtask timesteps $t[i]{=}t_j$ (darker cells indicate larger $t$).
  Middle (training): conditioned on $c\!=\!f_{\text{VLA}}(o,u)$, the diffusion head predicts $\epsilon_\theta(x_t[i],c,t[i])$, recovers $\hat{x}_0[i]$, and is supervised by the DF loss (noise MSE + end-of-trajectory and stage-end geometric terms + smoothness) \cite{chi2023diffusion,chen2024diffusion}.
  Right (inference): we \emph{partially denoise} a selected segment $S_{k,m}$ using jumpy DDIM \cite{song2020denoising} with geometry guidance $w\nabla_{x_t}U(\hat{x}_0;\mathcal{G})$; only the segment evolves while the complement is frozen, and the first segment is committed before replanning.}
  \label{fig:df}
\end{figure*}

\subsection{Problem Setup and Notation}
We consider language-conditioned, long-horizon manipulation where the agent must output a continuous action sequence $\vx_{1:L}\in\RR^{L\times d_a}$ (we use $d_a{=}7$) that satisfies intermediate \emph{stage} constraints and a final goal.
Let $u$ be the instruction, $\mathbf{o}$ the visual observation(s), and $\vc=f_{\text{VLA}}(\mathbf{o},u)\in\RR^{d_c}$ the embedding from a frozen or LoRA-tuned Octo or OpenVLA encoder \cite{team2024octo, kim2024openvla}.
Each trajectory optionally contains a stage-end mask $\mathbf{m}\in\{0,1\}^L$ and stage goals $\mathbf{G}\in\RR^{L\times 3}$ (EE geometry at stage ends), with $\mathcal{S}=\{i~|~m_i=1\}$, and a final positional goal $\mathbf{g}_{\text{final}}\in\RR^3$. We denote by $t_{\mathrm{now}}$ the index of the last executed token (prefix), and by $\hat{\vx}_0^{(i)}$ the clean estimate truncated at step $i$; see Table~\ref{tab:notation}.

\begin{table}[t]
\centering
\small
\caption{Notation used in the paper.}
\label{tab:notation}
\setlength{\tabcolsep}{6pt}               % 列间距，可调小到 4–5pt
\renewcommand{\arraystretch}{1.05}        % 行距，略紧
\begin{tabularx}{\columnwidth}{@{} l >{\raggedright\arraybackslash}X @{}}
\toprule
\textbf{Symbol} & \textbf{Meaning} \\
\midrule
$\vx_{1:L}$, $\vx_i\in\RR^{d_a}$ & Action tokens (position/orientation/gripper) \\
$\vc\in\RR^{d_c}$ & VLA embedding (Octo or OpenVLA) \\
$t[i]\in\{1,\dots,T\}$ & Per-token diffusion timestep \\
$\alpha_t$, $\baralpha_t$ & Noise schedule and cumulative product \\
$\veps$, $\veps_\theta$ & Gaussian noise or predicted noise \\
$\hat{\vx}_0$ & Predicted clean actions \\
$\mathcal{S}$, $\mathbf{G}$, $\mathbf{g}_{\text{final}}$ & Stage indices, stage goals, final goal \\
$\mathcal{G}=(\mathcal{V},\mathcal{E})$ & Scene graph (objects/relations) \\
$a=(k,m,s,w,\tau)$ & Meta action (start/length, stride, guidance, temperature) \\
$N(n),\,W(n,a),\,Q(n,a)$ & Visit count, cumulative return, mean action value at $n$ \\
$r_{\text{fast}}(n,a)$ & Fast reward for unexpanded/evaluable children \\
$S_{k,m}$ & Segment operator selecting tokens $[k:k{+}m{-}1]$ \\
\bottomrule
\end{tabularx}
\end{table}

\subsection{VLA-Conditioned Diffusion Head}
We parameterize a Transformer+FiLM diffusion head $f_\theta$ that predicts per-token noise conditioned on $\vc$.
For token $i$ at timestep $t[i]$,
\begin{equation}
    \veps_\theta\big(\vx_t[i],\,\vc,\,t[i]\big)\in\RR^{d_a}.
\end{equation}
FiLM derives token-wise scales/shifts/gates from $\vc$ (and optionally stage or subtask embeddings) to modulate each block \cite{chi2023diffusion}. Following variance-preserving diffusion \cite{ho2020denoising}, for each token $i$:
\begin{equation}
\label{eq:forward-token}
    \vx_t[i] \;=\; \sqrt{\baralpha_{t[i]}}\,\vx_0[i]
    \;+\; \sqrt{1-\baralpha_{t[i]}}\,\veps[i], 
    \qquad \veps[i]\sim\mathcal{N}(\mathbf{0},\vI),
\end{equation}
and the clean estimator is
\begin{equation}
\label{eq:x0hat-token}
    \hat{\vx}_0[i]
    \;=\;
    \frac{1}{\sqrt{\baralpha_{t[i]}}}
    \Big(\vx_t[i]-\sqrt{1-\baralpha_{t[i]}}\,\veps_\theta(\vx_t[i],\vc,t[i])\Big).
\end{equation}
Eqs.~\eqref{eq:forward-token}--\eqref{eq:x0hat-token} apply independently across tokens, enabling non-uniform noise levels across the sequence (``noise~=~mask'') \cite{chen2024diffusion}.

\subsection{Diffusion Forcing (DF) Objective}

We align the denoising schedule with subtask structure instead of treating all tokens equally. Given an ECoT-style stage mask $\mathbf{m}\!\in\!\{0,1\}^L$ and boundaries $\mathcal{S}\!=\!\{i:\,m_i\!=\!1\}$, we assign a \emph{single} timestep $t_j$ to all tokens within subtask $j$ (``noise\,$=$\,mask''), i.e., $t[i]{=}t_j$ if $i$ lies in subtask $j$ \cite{chen2024diffusion}. This yields clean predictions $\hat{\vx}_0[i]$ via (III.2)–(III.3). We then supervise with a stage- and geometry-aware objective:
\begin{align}
\mathcal{L}_{\text{DF}}
&= \underbrace{\tfrac{1}{L}\sum_{i=1}^{L}\!\big\|\veps_\theta(\vx_t[i],\vc,t[i])-\veps[i]\big\|_2^2}_{\text{\footnotesize noise MSE}} \nonumber\\
&\quad + \lambda_{\text{EoT}}\big\|\pos(\hat{\vx}_0^{(L)})-\mathbf{g}_{\text{final}}\big\|_2^2 \nonumber\\
&\quad + \lambda_{\text{stage}}\!\sum_{i\in\mathcal{S}}\big\|\pos(\hat{\vx}_0^{(i)})-\mathbf{G}_i\big\|_2^2 \nonumber\\
&\quad + \lambda_{\text{traj}}\tfrac{1}{L}\sum_{i=1}^{L}\big\|\hat{\vx}_0^{(i)}-\vx_0^{(i)}\big\|_2^2 \nonumber\\
&\quad + \lambda_{\text{smooth}}\tfrac{1}{L}\sum_{i=2}^{L}\big\|\Delta \hat{\vx}_0^{(i)}\big\|_2^2 .
\label{eq:df}
\end{align} 
\textbf{Implementation.}  We use a VP schedule with $T{=}1000$ steps and sample a single per-subtask timestep $t_j\!\sim\!\mathcal{U}[400,900]$ (``noise\,=\,mask''). This range balances denoising difficulty across stages and worked robustly across all suites.

\subsection{Partial Denoising with a Segment Operator}
At test time we only evolve a future segment while freezing the executed prefix. Instead of a diagonal matrix, we use a binary \emph{mask vector} $s\!\in\!\{0,1\}^L$ for the chosen segment $[k:k{+}m{-}1]$:
\[
s_i=\mathbb{1}[k\le i\le k{+}m{-}1],\quad \bar{s}= \mathbf{1}-s.
\]
With a schedule $t_{J-1}{=}\max(0,\,t_J{-}s)$ and temperature $\tau$
\textbf{Meta-parameters searched by MCTD.} segment start $k{\in}\{t_{\text{now}}{+}1,\dots\}$, length $m{\in}\{8,12,16\}$, stride $s{\in}\{2,4,8\}$, guidance $w{\in}[0,5]$, temperature $\tau{\in}[0.5,1.0]$. We restrict $k$ to start within $16$ tokens before the next stage boundary to focus search near subgoal interfaces.

\subsubsection{Geometry Relation-Aware Guidance}
We define a potential $U(\hat{\vx}_0;\mathcal{G})$ as a non-negative weighted sum of terminal alignment, stage anchoring, relation violation penalties from the scene graph, and collision costs; lower values indicate candidates that precisely land subgoals, satisfy symbolic relations, and remain collision-free. We form a guided noise with a practical Jacobian that ignores cross-token coupling:
\[
\tilde{\veps} \;=\; \veps_\theta(\vx_t,\vc,t)\;-\;w\;\underbrace{\Big(\mathrm{diag}(1/\sqrt{\bar\alpha_{t[i]}})\Big)^\top}_{\text{stop-grad on }\veps_\theta}\nabla_{\hat{\vx}_0}U,
\]
and substitute $\tilde{\veps}$ into the partial DDIM update (\ref{eq:forward-token}). In practice we compute $\nabla_{\hat{\vx}_0}U$ with autograd on the differentiable predicates (next paragraph) and cache it across the $m$ active tokens for the current jump.

\subsection{Scene Graph and Differentiable Predicates}
We maintain $\mathcal{G}=(\mathcal{V},\mathcal{E})$ via detection and VLM parsing \cite{li2024scene}.
Each relation $r\in\mathcal{R}$ induces a differentiable predicate $\phi_r$ (for guidance) and a satisfaction score $\mathrm{sat}_r\in[0,1]$ (for reward), e.g.,
\begin{equation}
\label{eq:phi-ontop}
\begin{aligned}
\phi_{\text{on-top}}(A,B) \;=\;
&\Big[\max\!\big(0,\,\delta_\perp-\|p_A^\perp-p_B^\perp\|\big)\Big]^2 \\
&\;+\;
\Big[\max\!\big(0,\,h_{\min}-(p_A^z-p_B^z)\big)\Big]^2, \\
\mathrm{sat}_{\text{on-top}} \;=\;&\;
\sigma\!\big(-\gamma\,\phi_{\text{on-top}}(A,B)\big).
\end{aligned}
\end{equation}
Other relations (\emph{in/left-of/aligned-with}) follow analogous distance or angle forms; collisions use SDF or capsule distances.
A stage-level textual parser (Qwen2.5-VL) extracts candidate relations (in, on top of, left-of, aligned) and anchors. For guidance and reward, each relation maps to a differentiable predicate based on signed distances and angular offsets. Collision penalties use capsule–capsule distances with safety margins (EE radius $r_{\mathrm{EE}}$, object radii $r_o$). 

\begin{figure}[t]
  \centering
  \includegraphics[width=\linewidth]{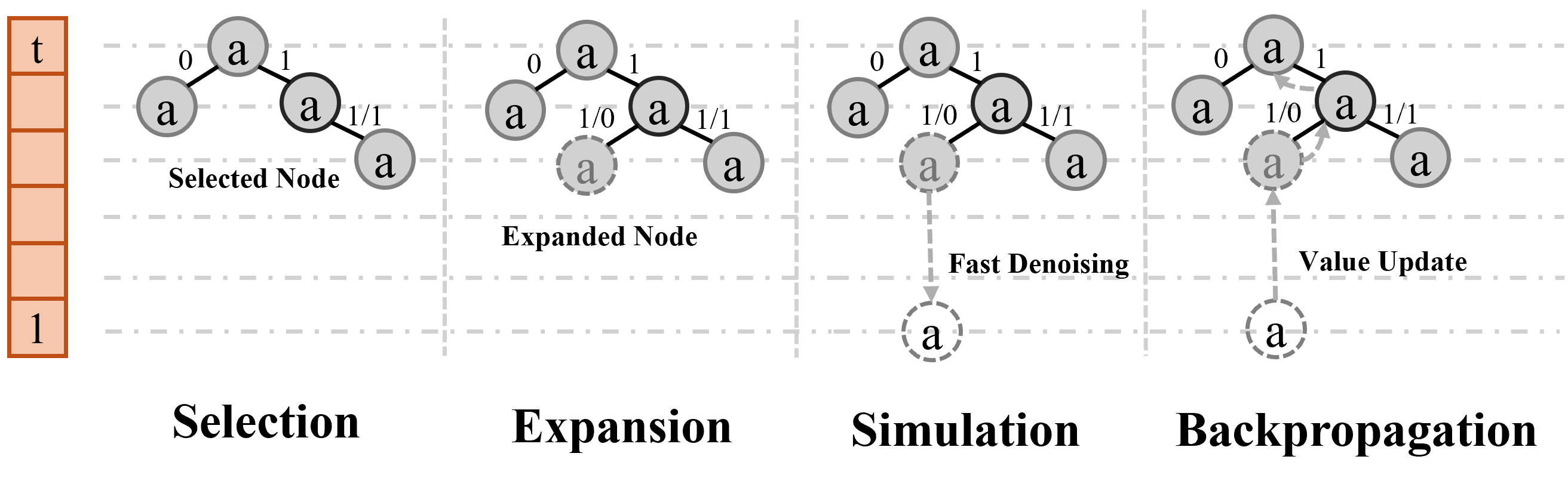}
  \vspace{-2mm}
  \caption{\textbf{Stage-aware MCTD.}
  For each stage we run an MCTS whose edges are meta-actions $a{=}(k,m,s,w,\tau,t)$.
  Selection uses P-UCT with scene-graph priors, \cite{silver2017mastering},
  and unvisited edges are ordered by a fast heuristic $r_{\text{fast}}(n,a)$.
  Simulation performs \emph{partial denoising} on $S_{k,m}$ (jumpy DDIM, guidance strength $w$) to produce a candidate $\hat{x}_0$, which is scored by the geometry-aware return $R(\hat{x}_0;\mathcal{G})$.
  Returns are backed up to update $(N,W,Q)$; the best child is executed for one segment, observations update $\mathcal{G}$, and the process recedes to the next stage.}
  \label{fig:mctd}
\end{figure}

\subsection{Stage-Aware MCTD with Dual Rewards}
\label{sec:stage-mctd}
\paragraph{Dual rewards: fast heuristic vs.\ true geometry return.}
We use two complementary signals. The \emph{fast} heuristic scores a child before we spend rollout compute: it combines (i) a language-consistent likelihood from the VLA (token-average log-probability of the meta-action, capturing whether the segment choice matches the instruction and visual context), (ii) a scene-graph prior that prefers segments which end near stage boundaries or make measurable progress toward unmet relations (e.g., moving a can toward the basket opening), and (iii) a cheap collision proxy based on minimum capsule or SDF margin along a straight-line waypoint proposal. This fast score is normalized, lightweight, and used to order unvisited children and to define the sampling distribution during simulation.

After we actually \emph{partially denoise} a segment and obtain a concrete candidate, we compute the \emph{true} geometry return on that candidate. It aggregates: terminal pose accuracy (position/orientation at the segment end or episode end), relation satisfaction from the scene graph (higher is better), collision penalties (lower is better), and a mild smoothness term around the edited boundary. All terms are scaled to comparable ranges and combined with fixed validation-tuned weights into a single scalar. The true return is the value that is backed up along the path to update $(N,W,Q)$ in MCTD. In short, the fast score guides exploration cheaply, while the true return provides faithful evaluation once a candidate has been generated.

\paragraph{Robust P-UCT selection.}
We select actions with Laplace smoothing to avoid $\log(0)$ and division by zero:
\begin{equation}
\label{eq:puct-robust}
\mathrm{Score}(n,a)
= Q(n,a) + c_{\text{puct}}\;P(n,a)\,
\sqrt{\frac{\log\!\big(N(n)+1\big)}{\,N(n,a)+1\,}},
\end{equation}
where $P(n,a)\propto \exp\{\psi(a;\mathcal{G})\}$.
If there exist unvisited children $\mathcal{U}=\{a: N(n,a)=0\}$, we pick among them by the fast reward; otherwise we use Eq.~\eqref{eq:puct-robust}:
\begin{equation}
\label{eq:select-rule}
a^\star =
\begin{cases}
\arg\max\limits_{a\in\mathcal{U}} r_{\text{fast}}(n,a), & \text{if } \mathcal{U}\neq\emptyset,\\[3pt]
\arg\max\limits_{a\in\mathcal{A}(n)} \mathrm{Score}(n,a), & \text{otherwise.}
\end{cases}
\end{equation}

\paragraph{Simulation with normalized sampling.}
During rollout, when a \emph{sampling} policy is enabled, we pick children by a softmax over fast rewards:
\begin{equation}
\label{eq:softmax-fast}
\begin{aligned}
p(a\mid n)
&=\frac{\exp\!\big(r_{\text{fast}}(n,a)-\max_{a'} r_{\text{fast}}(n,a')\big)}
        {\sum_{a''}\exp\!\big(r_{\text{fast}}(n,a'')-\max_{a'} r_{\text{fast}}(n,a')\big)}.
\end{aligned}
\end{equation}
which is well-defined even when all fast rewards are equal; a greedy policy is recovered by $\arg\max_a r_{\text{fast}}(n,a)$.

\begin{algorithm}[t]
\caption{PartialDenoise($\vx_{t_K}$, $a=(k,m,s,w,\tau)$, $f_\theta$,$\vc$,$\mathcal{G}$)}
\label{alg:partial}
\begin{algorithmic}[1]
\State Build grid $\{t_J\}_{J=K}^{0}$ with stride $s$; set $S\!\leftarrow\!S_{k,m}$, $\bar S\!\leftarrow\! I{-}S$
\For{$J=K,K\!-\!1,\dots,1$}
  \State Compute $\veps_\theta$ and $\hat{\vx}_0$ by \eqref{eq:x0hat-token}
  \State $\tilde{\veps}\leftarrow \veps_\theta - w\,\nabla_{\vx_{t_J}}U(\hat{\vx}_0;\mathcal{G})$ 
  \State Sample $\veta\sim\mathcal{N}(0,\tau^2 \vI)$
  \State $\vx_{t_{J-1}}\leftarrow \bar S\,\vx_{t_J} + S\big(\sqrt{\baralpha_{t_{J-1}}}\,\hat{\vx}_0 + \sqrt{1-\baralpha_{t_{J-1}}}\,\veta\big)$ 
\EndFor
\State \Return $\hat{\vx}_0$ from the last iteration
\end{algorithmic}
\end{algorithm}

\begin{algorithm}[t]
\caption{Stage-Aware MCTD with Dual Rewards}
\label{alg:stage-mctd}
\begin{algorithmic}[1]
\Require Stages $\{\mathsf{stage}_1,\dots,\mathsf{stage}_S\}$, VLA $f_{\text{VLA}}$, diffusion head $f_\theta$, budget per stage $B$
\State $\vc\!\leftarrow\! f_{\text{VLA}}(\mathbf{o},u)$; $\mathcal{G}\!\leftarrow\!\text{BuildSceneGraph}(\mathbf{o},u)$
\State Initialize root $n_0$ with current prefix and $\vx_{t_K}$; $n_0.\text{children}\leftarrow[\,]$
\For{$s=1$ \textbf{to} $S$} \Comment{stage-wise planning}
  \State Create stage node $n^{(s)}$ (parent $n_0$) and append to $n_0.\text{children}$
  \For{$b=1$ \textbf{to} $B$}
    \State $n\leftarrow n^{(s)}$, store path $\mathcal{P}\!\leftarrow\![\,]$
    \While{$n$ expandable}
      \State Generate $\mathcal{A}(n)$ with scene-graph prior $P(\cdot\mid n)$
      \State \textbf{if} $\exists a\in\mathcal{A}(n)$ with $N(n,a){=}0$ \textbf{then} $a^\star\!\leftarrow\!\arg\max_{a: N(n,a)=0} r_{\text{fast}}(n,a)$
      \State \textbf{else} $a^\star\!\leftarrow\!\arg\max_{a\in\mathcal{A}(n)} \mathrm{Score}(n,a)$ using Eq.~\eqref{eq:puct-robust}
      \State Append $(n,a^\star)$ to $\mathcal{P}$ and move to child $n\!\leftarrow\!\text{Child}(n,a^\star)$ (create if absent)
    \EndWhile
    \State $\hat{\vx}_0\!\leftarrow\!\textsc{PartialDenoise}(\vx_{t_K},a^\star,f_\theta,\vc,\mathcal{G})$
    \State $R\!\leftarrow\! R(\hat{\vx}_0;\mathcal{G})$
    \State \textbf{for} $(n,a)$ \textbf{in} $\mathcal{P}$ \textbf{do} update $(N,W,Q)$
  \EndFor
  \State Choose best child of $n_0$ for stage $s$ by visits or value; commit its first segment; observe, update $\mathcal{G}$
\EndFor
\State \Return concatenated $(\text{tasks},\text{states},\text{actions})$ trace for all stages
\end{algorithmic}
\end{algorithm}

\section{Experiments}

\begin{table*}[t]
\centering
\small
\renewcommand{\arraystretch}{0.9} % 更紧凑
\setlength{\tabcolsep}{5pt}
\caption{}
\label{tab:libero_sr_forge}
% \resizebox{\textwidth}{!}{
\begin{tabular}{lccccc}
\toprule
\rowcolor{headergray}
\multirow{2}{*}{Model} & \multicolumn{4}{c}{SR (\%)} & \multirow{2}{*}{Average} \\
\cmidrule(lr){2-5}
\rowcolor{headergray}
 & Libero-Spatial & Libero-Object & Libero-Goal & Libero-Long & \\
\midrule
\multicolumn{6}{l}{\emph{From scratch}}\\
Diffusion Policy \cite{chi2023diffusion} & 78.3 & 92.5 & 68.3 & 50.5 & 72.4 \\
MDT \cite{reuss2024multimodal}           & 78.5 & 87.5 & 73.5 & 64.8 & 76.1 \\
Seer (scratch) \cite{tian2024predictive}  & --   & --   & --   & 78.7 & --   \\
\midrule
\multicolumn{6}{l}{\emph{Fine-tuned from pretrained VLMs}}\\
Seer (fine-tuned) \cite{tian2024predictive}             & --   & --   & --   & 87.7 & --   \\
Dita / DiT Policy \cite{hou2025dita}               & 84.2 & 96.3 & 85.4 & 63.8 & 82.4 \\
TraceVLA \cite{zheng2024tracevla}                   & 84.6 & 85.2 & 75.1 & 54.1 & 74.8 \\
SpatialVLA \cite{qu2025spatialvla}                       & 88.2 & 89.9 & 78.6 & 55.5 & 78.1 \\
$\pi_{0}$ \cite{black2024pi_0}                     & 96.8 & \best{98.8} & \secondbest{95.8} & 85.2 & 94.2 \\
GR00T\text{-}N1 \cite{bjorck2025gr00t}             & 94.4 & 97.6 & 93.0 & 90.6 & 93.9 \\
Discrete Diffusion VLA                             & \best{97.2} & \secondbest{98.6} & \best{97.4} & \best{92.0} & \best{96.3} \\
OpenVLA \cite{kim2024openvla}                       & 84.7 & 88.4 & 79.2 & 53.7 & 76.5 \\
Octo-Base \cite{team2024octo}                   & 78.9 & 85.7 & 84.6 & 51.1 & 75.1 \\
\midrule
\rowcolor{oursrow}
\textbf{FORGE-Tree + OpenVLA} & \textbf{92.4} & \textbf{94.7} & \textbf{89.3} & \textbf{83.2} & \textbf{89.9} \\
\rowcolor{oursrow}
\textbf{FORGE-Tree + Octo}    & \textbf{96.6} & \textbf{88.2} & \textbf{93.4} & \textbf{91.0} & \textbf{92.3} \\
\bottomrule
\end{tabular}
% }
\end{table*}

\begin{figure}[!t]
    \centering
    \includegraphics[width=1\linewidth]{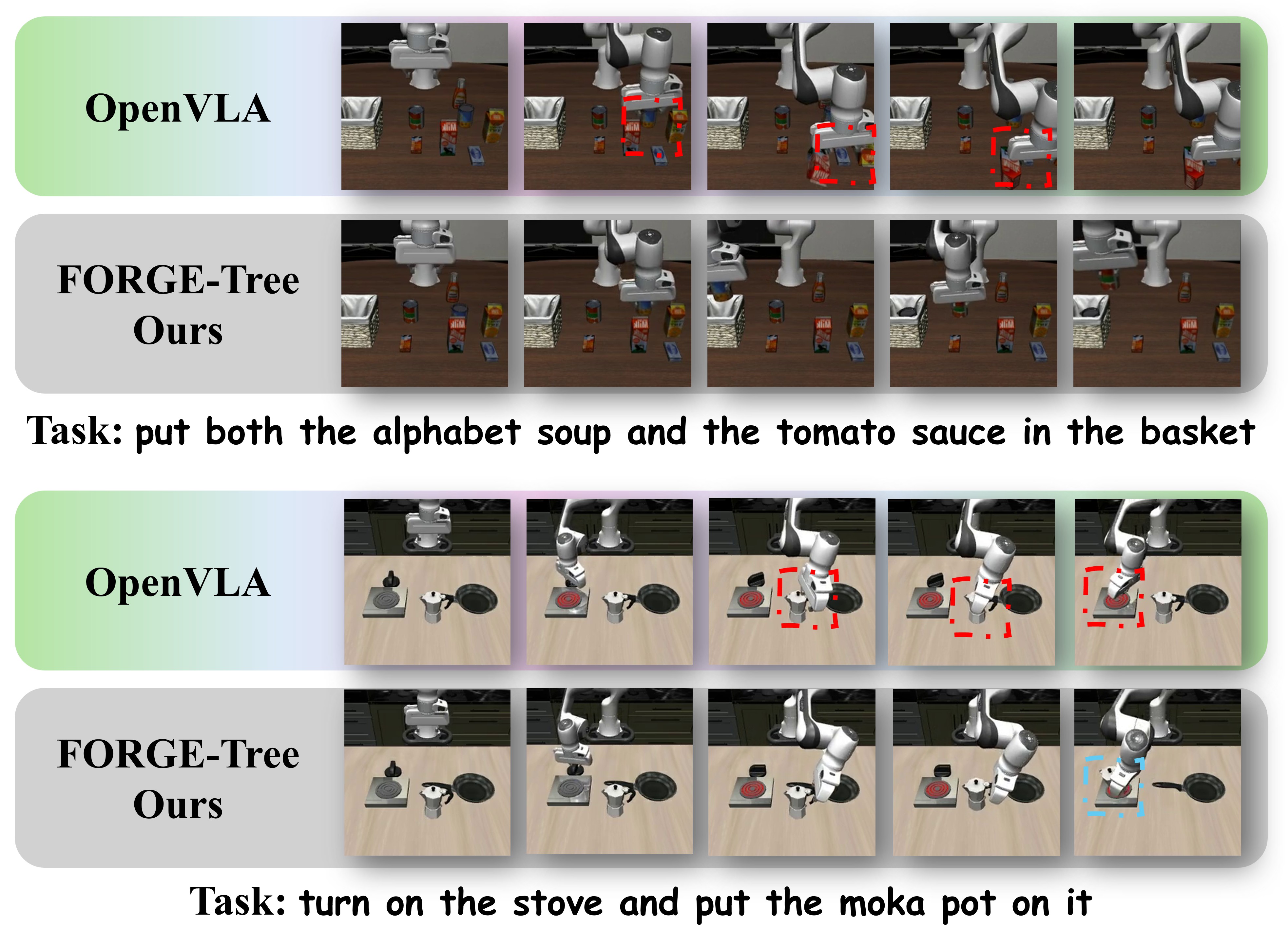}
    \caption{Top: OpenVLA collides (red dashed boxes). Bottom: \textbf{FORGE-Tree} edits only the upcoming segment with geometry guidance and places actions inside the goal region with proper clearance. }
    \label{fig:Geometry-aware partial denoising}
\end{figure}

\subsection{Benchmarks and Baselines}
\paragraph{\textbf{Benchmarks.}} We train on \textbf{EMMA-X} (stage-annotated manipulation trajectories) \cite{sun2024emma} and evaluate in simulation on LIBERO-Spatial, LIBERO-Object, LIBERO-Goal, and LIBERO-Long (10 tasks per suite; 500 expert demos per suite) and ManiSkill (diverse physics tasks)  following each suite's official protocol (episode budgets, success definitions, and resets) \cite{liu2023libero, mu2021maniskill}. Stage boundaries come from an ECoT-style parser; we also extract a \emph{Qwen2.5-VL} \cite{bai2025qwen2} embedding as additional context for stage conditioning and scene-graph construction. Stage parsing is used only to align training targets on EMMA-X; at test time, no stage labels are used, the search selects segments from decoded tokens alone. Both the parser and the Qwen2.5-VL encoder are frozen and are not fine-tuned on LIBERO or ManiSkill. All methods use the same observations (RGB only unless noted), action space (7-DoF EE pose + gripper), and demonstration splits. We train \emph{only} the control head on EMMA-X and \emph{do not} fine-tune on LIBERO or ManiSkill. All results are \emph{zero-shot evaluations} of the control layer on top of frozen VLAs (OpenVLA and Octo). Baselines that report numbers using LIBERO or ManiSkill demonstrations are taken from their official protocols or re-runs; we do \emph{not} use these demonstrations to train our head.
\begin{figure*}[!t]
    \centering
    \includegraphics[width=0.85\linewidth]{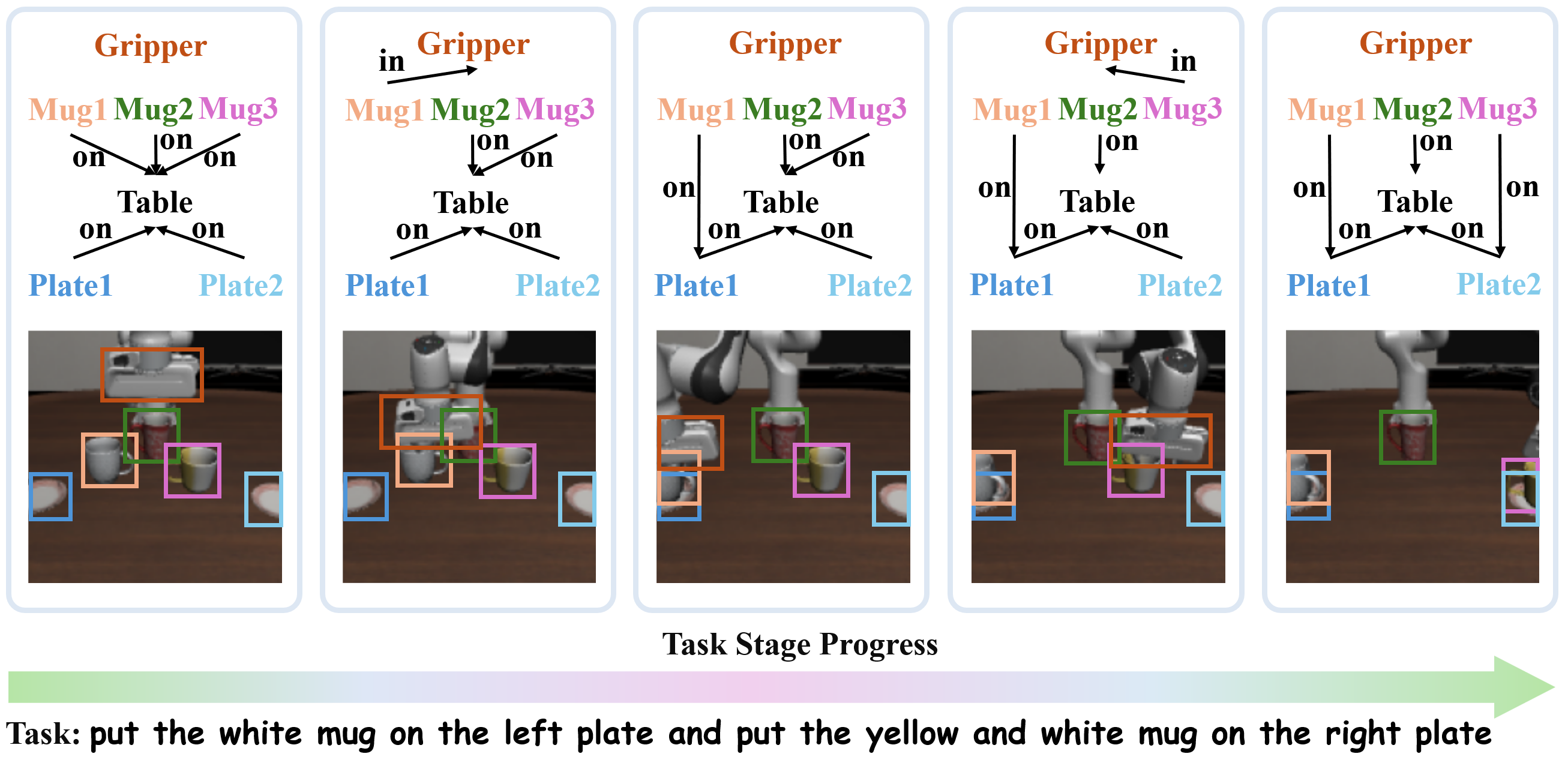}
    \caption{Left→right: snapshots from a LIBERO task (“put the white mug on the left plate and put the yellow mug on the right plate”) with \textbf{FORGE-Tree}. In each panel, the top diagram shows the current scene graph $\mathcal{G}$ with nodes (Gripper, Mug1–3, Plate1–2, Table) and edges labeled by relations (\texttt{in}, \texttt{on}); the bottom image shows the corresponding RGB frame with color-matched boxes. As the episode advances, relations update (e.g., \texttt{in}(Gripper, Mug), \texttt{on}(Mug, Plate)), marking subgoal completion. FORGE-Tree uses this graph both to bias expansion (priors over meta-actions) and to score candidates via geometry relation predicates, enabling stage-wise planning that satisfies the relational goals.}
    \label{fig:Scene graph stage progression }
\end{figure*}
\paragraph{\textbf{Backbones.}} Unless otherwise stated, the Vision–Language-Action encoder is \emph{frozen} OpenVLA or Octo; optional LoRA uses the official OpenVLA implementation. Our controller is the proposed VLA-conditioned diffusion head trained with Diffusion Forcing (DF) and executed with MCTD at test time.

\paragraph{\textbf{Baselines.}} We compare against representative policies spanning the two dominant paradigms—\emph{autoregressive (AR) token decoders} and \emph{continuous diffusion or flow-matching} policies—covering both models trained from scratch and fine-tuned from large pretrained bases. 
\textbf{AR action decoders.} RT-1-X / RT-2-X~\cite{o2024open}, OpenVLA~\cite{kim2024openvla}, Octo-Small / Octo-Base~\cite{team2024octo}, HPT~\cite{wang2024scaling}, TraceVLA~\cite{zheng2024tracevla}, and SpatialVLA~\cite{qu2025spatialvla} instantiate AR-style generation of discrete action tokens on a unified VLM backbone.
\textbf{Continuous diffusion or flow-matching.} Diffusion Policy~\cite{chi2023diffusion}, MDT~\cite{reuss2024multimodal}, DiT Policy (DiTA)~\cite{hou2025dita}, $\pi0$~\cite{black2024pi_0}, and GR00T-N1~\cite{bjorck2025gr00t} implement denoising or flow-matching heads over continuous action trajectories; Seer~\cite{tian2024predictive} is reported in both scratch and fine-tuned forms. 

\textit{Control-layer comparison.} Because our method is a \emph{plug-in control layer} over frozen VLA backbones, we additionally report \emph{OpenVLA and Octo + FORGE-Tree} (DF-only and DF+MCTD) alongside native baselines to highlight the incremental gains achievable without changing the encoder. Numbers marked “+ FORGE-Tree” reuse the exact frozen checkpoint and differ only in the control layer at inference.

\subsection{Performance Comparisons}
\paragraph{\textbf{Training details.}} We train \emph{only} the diffusion control head with Diffusion Forcing (DF) on top of a frozen VLA encoder; unless stated, there is no benchmark-specific fine-tuning on LIBERO or ManiSkill. Inputs follow the backbone’s preprocessing (RGB $224{\times}224$, backbone normalization). Actions are 7-D (EE $x,y,z$, orientation, gripper), standardized per-dataset and clipped to $[-3\sigma,3\sigma]$. From EMMA-X we sample windows of $L{=}64$ actions (pad/trunc as needed) with ECoT-derived stage boundaries. For each window, we draw $1$–$2$ stages and assign a constant per-stage timestep $t_j\!\sim\!\mathcal{U}[400,900]$ (others $t{=}0$); diffusion horizon $T{=}1000$ with a cosine schedule. The DF objective uses fixed weights $\lambda_{\text{EoT}}{=}5.0$, $\lambda_{\text{stage}}{=}3.0$, $\lambda_{\text{traj}}{=}1.0$, $\lambda_{\text{smooth}}{=}0.1$; padding is fully masked. Optimization: AdamW ($\beta{=}(0.9,0.999)$, weight decay $0.05$), peak LR $3{\times}10^{-4}$ (control head) with cosine decay and $2$k warmup, total $200$k steps, global-norm clipping $1.0$, EMA $0.999$. Training uses bf16 and PyTorch DDP on 4$\times$48\,GB Ada GPUs, per-GPU batch $64$ with $\times 2$ accumulation (effective $512$). Augmentations are mild: color jitter $0.2$, horizontal flip $p{=}0.5$ when mirror-invariant, and Gaussian action noise $\mathcal{N}(0,0.01)$; we avoid heavy geometric transforms to preserve metric geometry.

\paragraph{\textbf{LIBERO results.}} Table~\ref{tab:libero_sr_forge} reports success rates (SR) on the four LIBERO suites. As a \emph{control-layer} add-on, \textbf{FORGE-Tree} delivers substantial gains over its underlying VLAs without changing the encoder: on \textbf{OpenVLA} it lifts the average SR from $76.5\%$ to $89.9\%$ (\textbf{+13.4 pp}), with per-suite improvements of \textbf{+7.7} (Spatial), \textbf{+6.3} (Object), \textbf{+10.1} (Goal), and \textbf{+29.5} pp (Long). On \textbf{Octo}, it raises the average from $75.1\%$ to $92.3\%$ (\textbf{+17.2 pp}), with suite-wise gains of \textbf{+17.7}, \textbf{+2.5}, \textbf{+8.8}, and \textbf{+39.9} pp, respectively. The largest improvements occur on \emph{LIBERO-Long}, consistent with our stage-wise partial denoising and geometry-aware planning designed for multi-step, contact-rich sequences. 

For context, state-of-the-art discrete-diffusion decoders report an average SR of $96.3\%$ on these suites; FORGE-Tree \emph{narrows the gap to within} \textbf{4--6 pp} while operating as a plug-in control layer on frozen OpenVLA or Octo backbones. Compared with from scratch diffusion or flow-matching policies, our results remain competitive across suites, highlighting that test-time, tree-structured denoising can recover a large fraction of long-horizon performance without retraining the vision–language encoder.

\subsection{Ablation Study}
We compare the full model against (i) \emph{no stage guidance} in DF and (ii) \emph{no scene graph} in MCTD. Removing stage guidance consistently lowers SR and increases terminal pose error, most notably on LIBERO–Goal and Long, which decrease by \textbf{14\%} and\textbf{ 27\%} average each, showing that stage-aligned supervision is essential for landing intermediate subgoals. Removing the scene graph reduces relation satisfaction, increases collisions, and requires larger search budgets to match SR, demonstrating that $\mathcal{G}$ supplies both effective exploration priors and geometry-aware evaluation. These results confirm that stage-aware DF and scene-graph structure are key to long-horizon reliability.

\begin{table}[t]
  \centering
  \small
  \setlength{\tabcolsep}{4.2pt}
  \caption{Effect of removing stage guidance (OpenVLA). Absolute SR and relative change vs.\ Full.}
  \label{tab:ablate-stage-compact}
  \begin{tabular}{l S[table-format=2.1] S[table-format=-2.0] S[table-format=2.1] S[table-format=-2.0]}
    \toprule
    & \multicolumn{2}{c}{\textbf{LIBERO–Goal}} & \multicolumn{2}{c}{\textbf{LIBERO–Long}} \\
    \cmidrule(lr){2-3}\cmidrule(lr){4-5}
    \textbf{Variant} & \textbf{SR (\%)} & $\boldsymbol{\Delta}$\textbf{SR} & \textbf{SR (\%)} & $\boldsymbol{\Delta}$\textbf{SR} \\
    \midrule
    \textbf{Full (ours)}           & {\bf [88.2]} & {0} & {\bf [83.2]} & {0} \\
    \textbf{w/o Stage guidance}     & {[75.9]}     & \textbf{-14} & {[60.7]}     & \textbf{-27} \\
    \bottomrule
  \end{tabular}
\end{table}

\subsection{Qualitative Failure Analysis}
We observed three recurring failures for \textsc{FORGE-Tree}. 
(i) \textit{Scene-graph errors.} False/weak detections (e.g., a shallow rim or occluded handle) inject noisy relations, 
so search refines a non-critical segment while the terminal segment remains suboptimal.  (ii) \textit{Search overhead and slow rewards.} Tree expansion can be compute-heavy when end-of-segment evaluation relies on expensive geometry checks, delaying value estimates and reducing effective breadth. (iii) \textit{Stall at the first grasp.} When the initial pick fails to establish stable contact, the policy keeps revisiting the same short segment with conservative edits, making little physical progress.

\section{CONCLUSION}
This work introduces\textbf{ FORGE-Tree}, a control-layer framework that couples a VLA-conditioned \emph{Diffusion Forcing} head with test-time \emph{Monte Carlo Tree Diffusion}. The key idea is to plan in the space of \emph{editable trajectory segments}: training aligns denoising with stage structure so the decoder learns to land intermediate subgoals, while inference performs \emph{tree-structured denoising} over meta-actions (segment, stride, guidance, temperature) guided by a scene graph. This turns diffusion decoding into a budget-scalable planner that preserves a good prefix and allocates compute where geometry is tight.

Our study suggests a practical path to stronger long-horizon manipulation by \emph{augmenting} rather than replacing pretrained VLAs. Looking forward, we aim to \underline{(i)} deploy on real hardware and study sim-to-real effects, \underline{(ii)} jointly learn scene-graph perception with task priors, \underline{(iii)} amortize meta-action proposals to reduce planning latency, and \underline{(iv)} extend to partially observed, multi-object settings with tighter contact modeling. We hope this bridges symbolic task structure and continuous diffusion control in a way that remains modular, interpretable, and compute-aware.

% \addtolength{\textheight}{-12cm}   % This command serves to balance the column lengths
                                  % on the last page of the document manually. It shortens
                                  % the textheight of the last page by a suitable amount.
                                  % This command does not take effect until the next page
                                  % so it should come on the page before the last. Make
                                  % sure that you do not shorten the textheight too much.

%%%%%%%%%%%%%%%%%%%%%%%%%%%%%%%%%%%%%%%%%%%%%%%%%%%%%%%%%%%%%%%%%%%%%%%%%%%%%%%%

%%%%%%%%%%%%%%%%%%%%%%%%%%%%%%%%%%%%%%%%%%%%%%%%%%%%%%%%%%%%%%%%%%%%%%%%%%%%%%%%

%%%%%%%%%%%%%%%%%%%%%%%%%%%%%%%%%%%%%%%%%%%%%%%%%%%%%%%%%%%%%%%%%%%%%%%%%%%%%%%%
% \section*{APPENDIX}

% Appendixes should appear before the acknowledgment.

%%%%%%%%%%%%%%%%%%%%%%%%%%%%%%%%%%%%%%%%%%%%%%%%%%%%%%%%%%%%%%%%%%%%%%%%%%%%%%%%

\bibliographystyle{IEEEtran} % 或 unsrtnat、IEEEtran 等
\bibliography{references}

\end{document}